# RIM Hand : A Robotic Hand with an Accurate Carpometacarpal Joint and Nitinol-Supported Skeletal Structure

Joon Lee[1],[†] Jeongyoon Han[1],[†] Doyoung Kim[1], *SeokhwanJeong*[1*]

[1]Department of Mechanical Engineering, Sogang University,

35, Baekbeom-ro, Mapo-gu, Seoul, Republic of Korea

E-mail: seokhwan@sogang.ac.kr

**Abstract:** This paper presents the flexible RIM Hand, a biomimetic robotic hand that precisely replicates the carpometacarpal (CMC) joints and employs superelastic Nitinol wires throughout its skeletal framework. By modeling the full carpal-to-metacarpal anatomy, the design enables realistic palm deformation through tendon-driven fingers while enhancing joint restoration and supports skeletal structure with Nitinol-based dorsal extensors. A flexible silicone skin further increases contact friction and contact area, enabling stable grasps for diverse objects. Experiments show that the palm can deform up to 28%, matching human hand flexibility, while achieving more than twice the payload capacity and three times the contact area compared to a rigid palm design. The RIM Hand thus offers improved dexterity, compliance, and anthropomorphism, making it promising for prosthetic and service-robot applications.

[†]These authors contributed equally to this work and are considered to be co-first authors.

# 1 Introduction

The human hand is one of the most versatile organs in the body. With more than 20 degrees of freedom (DOF), over 27 bones [1], and numerous muscles and nerves acting in coordination, the hand exhibits both flexibility and rigidity capable of performing delicate movements. Through its sophisticated and dexterous movements, it can interact widely with the surrounding environment [2].

Advances in science and technology have driven extensive research on robotic hands aimed at mimicking human-like motions and range of motion (ROM). However, designs are often limited by the need to minimize the number of actuators within constrained spaces. Consequently, many studies employ underactuated mechanisms, where a small number of actuators control multiple joints, or fixed-linkage structures to reduce extraneous motion and enhance control stability. These approaches seek to provide multi-DOF functionality using as few actuators as possible [3, 4, 5].

However, robotic hands primarily composed of rigid components often encounter challenges in reproducing the flexible and delicate motions of the human hand. As a result, they may not always provide the necessary ROM for grasping unstructured objects or conducting various daily tasks. Moreover, introducing multiple DOF typically requires numerous links and joints, which increases design complexity and may limit adaptability across diverse operational environments.

To overcome these limitations, research has focused on biomimetic design, integrating the structural and mechanical characteristics of the human body into robotic hands [6, 7]. For instance, tendon-driven mechanisms emulate muscle-tendon architectures for softer interactions, while other approaches replicate the flexibility and durability of bones and joints.

ACT Hand is a biomechanically designed robotic hand that precisely analyzes and incorporates the anatomical structures of human fingers and tendons. It applies a tendon-driven mechanism to emulate the delicate motions and force-transmission methods of the human hand [8]. In addition, various studies have proposed imitating ligaments and bones [9] or minimizing the number of actuators and improving efficiency[10, 11, 12, 13]. Meanwhile, efforts to mimic the shape and mechanics of the interphalangeal joints have included rolling-joint designs to provide robustness against off-axis external impacts [14], as well as synovial-joint-inspired models to enhance flexibility [15].

Nevertheless, most existing research has primarily focused on relatively simple distal phalangeal (DIP), proximal interphalangeal (PIP) and metacarpophalangeal(MCP) joint structures (see Fig. 2(a)). Few designs take into account the mechanics and structure spanning from the metacarpals in the palm to the wrist. Notably, the carpometacarpal (CMC) joints have fixed DOF that can improve grasp flexibility and adaptability [16], thereby enhancing overall grasp functionality. Implementing these joints in robotic hands can confer human-like motion and appearance, yet such designs have not been explored in depth.

Although some studies have proposed designs that consider the palm's DOF, they often involve a single hinge mechanism with only one DOF [17, 10], or high-DOF palms that do not sufficiently reproduce the ROM and DOF of the human hand [16]. Moreover, the palm's contribution to interacting with the external environment has not been fully studied.

In parallel, ongoing works aim to achieve more compact and lightweight robotic hands. Traditional approaches rely on mechanical components such as ball bearings and springs to implement joints. However, these designs require complex assembly processes and occupy substantial internal space, impeding size reduction. To address these issues, some have utilized 3D printing technology for compact hand structures [18], or minimized the total number of components to optimize space utilization [19].

Despite ongoing research, attaining a balance between mechanical stability, flexibility, and compactness presents difficulties. Nickel-titanium alloy (Nitinol) has emerged as a promising solution, offering notable restoring force and flexibility shown in medical and rehabilitation robotics [20, 21, 22, 23, 24]. However, few robotic hands employ Nitinol as the main structural material, and fully exploit it in comprehensive hand design.

In this study, we introduce an anthropomorphic robotic hand, termed the RIM Hand, that includes the full mechanical structure of the human fingers and palm. An overview of the proposed RIM Hand is provided in Fig. 1 and Video 1 (supplementary material). Our design closely emulates the overall skeletal framework of the human hand, from the carpal bones through the metacarpals, by implementing rolling joints for the phalanges and a drive mechanism inspired by human muscle actuation. In particular, Nitinol was used to provide restorative forces for the finger, thumb, and palm joints and to support the entire robotic hand frame. Additionally, to replicate the soft tissues of the human palm, a flexible skin was fabricated to achieve realistic palm deformation. Through these contributions, our work provides a comprehensive framework for biomimetic robotic hand hardware design, integrating both the structural and functional attributes of the human hand.

The main contributions of this paper are as follows:

1. A fully biomimetic palm structure that incorporates the complete motion and DOF of the metacarpals and CMC joints, achieving realistic palm deformation and robust grasps for a wide range of objects.

2. A compact Nitinol-based skeletal framework that supports skeletal structure and improves spatial efficiency through flexible dorsal restorations.

3. An anatomically inspired thenar design that closely reflects the shape and motion of the thumb muscles, enhancing natural movement, grasp stability, and visual anthropomorphism.

The remainder of this paper is organized as follows. Section II describes the structural design of the proposed robotic hand. Section III presents the finger kinematics. Section IV demonstrates the experimental validation. Finally, Section V concludes the paper.

# 2 Design and Mechanism

## 2.1 Design

In this section, we present the overall design of the robot hand by dividing it into three main components: the fingers, the palm, and the skin.

### 2.1.1 Finger Design

Designing a robot hand is challenging due to significant variation in individual hand shapes. However, the relative proportions among finger joints are generally consistent [25]. In this study, the average hand length of Korean adults in their 20s [26] was adopted as a reference to preserve these proportional relationships across finger segments. Incorporating anatomical dimensions into robot-hand design improves functional performance and human-like aesthetics. Table 1 presents the lengths and ratios of each finger segment for the proposed RIM Hand.

### 2.1.2 Palm Design

The palm consists of five CMC joints connecting the carpal bones to the metacarpals (see Fig. 2(a)). Each CMC joint provides specific DOF essential for robust grasp mechanics [27, 16]. The first CMC joint is a saddle joint with two DOF, enabling the wide range of thumb motion. The second and third are plane joints, offering limited translational movement and contributing overall hand stability [28]. In contrast, the fourth and fifth condyloid joints permit palmar rotation that facilitates palm deformation.

In the proposed RIM Hand, the palm structure imitates the metacarpals and carpal bones to capture distinct characteristics of each CMC joint (see Fig. 2(b)). All MCP, PIP, and DIP joints are 1-DOF rolling-joints for flexion. The first CMC joint for the thumb movement, which accounts for about 40% of overall hand function [29], uses a customized ball joint module with a spherical bearing and linear shaft, supporting 3-DOF and a thumb ROM up to 55° [30] (Fig. 2(c)). Further details on the RIM Hand's CMC joints appear in Table 2.

In addition, an arch structure was devised by referencing the shape of human carpal bones to facilitate an accurate grasp posture as illustrated in Fig. 2(d). Specifically, the fifth metacarpal in the human hand is rotated by approximately 10° to 15° toward the palmar plane, while the first metacarpal is rotated by about 40° to maintain equilibrium state. Based on these anatomical considerations, the neutral configuration of the RIM Hand's carpal bones was precisely designed.

### 2.1.3 Skin Design

Prior research has introduced skins to increase the contact area and grasp stability [31], protect internal components from impacts [32], and provide a human-like appearance [33]. However, most existing approaches rely on rigid metal or single-piece coverings [34], which limit the contact area and concentrate grip force in a smaller region, reducing overall grasp efficiency.

To overcome these limitations, the RIM Hand adopts flexible materials for both fingers and palm, thereby increasing friction and enlarging the contact area. As shown in Fig. 3, the finger and palm skins are separately fabricated. This closely matches human finger dimensions and enhances grasp friction [35].

Unlike previous designs that simply attach flexible skins to rigid frames or restrict palm motion to a single DOF [36], the RIM Hand enables at least five DOF in palm deformation (see Fig. 5(a)). The palm is molded from Ecoflex 00-30 silicone with internal air pocket that reduces stiffness and supports full range of thumb opposition. This configuration preserves natural thenar movement and markedly improves grasp stability. The skin fabrication process is illustrated in Fig. 3(b)-(e).

## 2.2 Mechanism

### 2.2.1 Tendon-driven Mechanism

The proposed RIM Hand adopts a tendon-driven mechanism to execute essential hand motions (e.g., power grip, pinching) with minimal actuators (servo motors). As shown in Fig. 4(a) and (b), the Flexion Tendon, mirroring the configuration of the Flexor Digitorum Superficialis (FDS), mainly handles PIP joint flexion, while the Lumbrical Tendon flexes the MCP joint. Additionally, a PIP-DIP Coupling Tendon (yellow line in Fig. 4(b)) ensures synchronized motion at a 3:2 ratio [15], enabling a single actuator to drive both PIP and DIP joints simultaneously.

The thumb plays a pivotal role in allowing various complex hand movements [29]. As shown in Fig. 4(a), the thenar muscle group, consisting of the Opponens pollicis, Adductor pollicis, and Abductor pollicis, forms the anatomical basis for key thumb movements such as opposition, adduction, and abduction. Based on this foundation, the thumb of the RIM Hand is actuated with five tendons to offer high DOF and natural motion. Two tendons (Flexion and Lumbrical) control flexion at MCP and interphalangeal (IP) joints, while the other three manage opposition, adduction, and abduction at the CMC joint (Fig. 4(c)). Referencing the structure of the thenar muscles, this design provides up to 3-DOF rotation of the thumb. As illustrated in Fig. 4(d), these tendons are integrated into the robotic hand.

#### 2.2.2 Joint Return Mechanism

The proposed robotic hand employs a Nitinol-based mechanism to passively restore finger posture after flexion. Nitinol wire, crimped with a sleeve, was positioned along the finger's center axis to serve as the return mechanism (Fig. 5(a)). For the PIP and DIP joints, a 0.584 mm Nitinol wire (red) is arranged in a fixed-free configuration. When the finger bends, the Nitinol wire slides at its free end and curves with the joint of bending radius, generating a restoring force that brings the finger back to neutral pose (Fig. 5(c), (d)). The MCP and the fourth, fifth CMC joints use two 0.584 mm wires (blue, purple) arranged in a fixed-free configuration. The first CMC joint is restored by two crossed Nitinol wires for multi-DOF motion (Fig. 5(b)).

According to the bending tests of Weaver et al. [40] and Yan et al. [41], all joints of the RIM Hand achieve fatigue lives exceeding 10,000 cycles and are within the elastic deformation region across the full operating range. Fig. 5(e) presents an ANSYS simulation that replicates the Nitinol behavior of the RIM Hand during actuation, evaluating stress as a function of the Nitinol rotation angle. This results also confirm that the Nitinol behaved within elastic deformation region. (Detailed values are given in Table 3 and Table 4)

Although inherently stable, small-scale rolling contact joints can experience slight misalignment under sliding and torsional loading, often resulting from tendon elongation or fabrication tolerances. The Nitinol wire located along the central axis additionally assists in mitigating such sensitivity. Overall, this mechanism supports the hand's structure while ensuring compact, robust restoration of each joint.

# 3 Kinematic Modeling

As shown in Fig. 6(a), the rolling-joint consists of two cylindrical elements in contact, where $\theta$ denotes the rotation angle of the line connecting their center, and $\theta$ represents the flexion angle of the upper phalanx relative to the lower phalanx. Assuming ideal rolling without slip, both cylindrical surfaces are designed with equal radii ($r_1 = r_2$), enabling consistent and predictable rotational motion. In this case, two types of radii are defined: the outer radii ($r_1$ and $r_2$), where rolling contact occurs, and the inner radii ($r_{r1}$ and $r_{r2}$), around which the joint coupling tendon is wound.

$$\frac{r_1}{r_2} = \frac{r_{r1}}{r_{r2}} \tag{1}$$

The ratio between the outer and inner radii is defined in (1), from which the joint angle can be derived once the radii are specified. Under this condition, the relationship between $\theta$ and $\theta_r$ is constrained by the rolling joint kinematic condition in (2), as described in [15].

$$\theta_r = \frac{r_2}{r_1 + r_2}\theta = \frac{1}{2}\theta \; (\because r_1 = r_2) \tag{2}$$

The overall finger kinematics model is illustrated in Fig. 6(b), where local coordinate frames are assigned from the finger base (frame 0) to the fingertip (frame 8). The Denavit-Hartenberg (DH) parameters including link lengths and joint angles are determined based on measured dimensions, actuation ranges, and kinematic constraints. As a representative case, the DH parameters for the middle finger are presented in Table 5, and the same modeling approach is applied to the other fingers. Here, $i = 0$ corresponds to the base of Link 1, aligning the frame indices with the link numbering in Table 5. The homogeneous transformation matrix from frame $i$ to $i + 1$ is denoted as ${}^{i}T_{i+1}$. By sequentially multiplying these matrices from frame 0 to 8, the overall finger pose is obtained, as shown in (3). The fingertip coordinates $P_x$, $P_y$, $P_z$ can then be expressed in closed form with the link parameters and joint angles, yielding the position equations in (4). The actuation range is determined by referencing the natural range of the human finger, from the carpal bones to each phalanx.

$$\begin{bmatrix} P_x \\ P_y \\ P_z \\ 1 \end{bmatrix} = {}^{0}T_{1}\,{}^{1}T_{2}\,{}^{2}T_{3}\,{}^{3}T_{4}\,{}^{4}T_{5}\,{}^{5}T_{6}\,{}^{6}T_{7}\,{}^{7}T_{8} \begin{bmatrix} 0 \\ 0 \\ 0 \\ 1 \end{bmatrix} \tag{3}$$

$$
\begin{aligned}
P_x &= 13 + 82.5\cos\left(\frac{\theta_{\mathrm{MCP}}}{2}\right) + 32.261\cos(\theta_{\mathrm{MCP}}) \\
&\quad + 12\cos\left(\theta_{\mathrm{MCP}} + \frac{\theta_{\mathrm{PIP}}}{2}\right) + 17.478\cos\left(\theta_{\mathrm{MCP}} + \theta_{\mathrm{PIP}}\right) \\
&\quad + 8\cos\left(\theta_{\mathrm{MCP}} + \frac{4}{3}\theta_{\mathrm{PIP}}\right) + 16.261\cos\left(\theta_{\mathrm{MCP}} + \frac{5}{3}\theta_{\mathrm{PIP}}\right), \\
P_y &= 82.5\sin\left(\frac{\theta_{\mathrm{MCP}}}{2}\right) + 32.261\sin(\theta_{\mathrm{MCP}}) \\
&\quad + 12\sin\left(\theta_{\mathrm{MCP}} + \frac{\theta_{\mathrm{PIP}}}{2}\right) + 17.478\sin\left(\theta_{\mathrm{MCP}} + \theta_{\mathrm{PIP}}\right) \\
&\quad + 8\sin\left(\theta_{\mathrm{MCP}} + \frac{4}{3}\theta_{\mathrm{PIP}}\right) + 16.261\sin\left(\theta_{\mathrm{MCP}} + \frac{5}{3}\theta_{\mathrm{PIP}}\right)
\end{aligned} \tag{4}
$$

The first CMC joint is modeled with three degrees of freedom by employing a spherical bearing that offers a maximum operational range of -55° to 55°. The DH parameters were derived based on the position vectors and kinematic analysis of the thumb's opposition, adduction, and abduction tendons discussed in Section II.

Based on data from [7], Table 6 compares the range of motion of the RIM Hand's finger joints with that of human hand and prosthetic hands. This comparative analysis based on angular characteristics demonstrates that the RIM Hand provides a sufficient range of motion for daily tasks.

To evaluate the kinematic control of the RIM Hand, line- tracking and point-tracking experiments were conducted (see Fig. 6(d)-(e)). As shown in Fig. 6(c), fingertip positions were manually marked and tracked in the software Tracker, with the baseline established using a level ruler placed parallel to the ground in the experimental setup. For each target point, the required joint angles were computed via inverse kinematics, and the position error was obtained by comparing the actual fingertip position of the RIM Hand with the target position. Each test was repeated ten times with identical target trajectories. In the line-tracking test, the hand achieved an average angular error of about 5°, indicating stable and accurate tracking performance. In the point-tracking test, a predefined triangular set of points was tracked, yielding a maximum absolute mean error of 3.78 mm and a mean repeatability error of 0.488 mm. In tendon-driven mechanisms, nonlinearities arising from tendon slack, friction along the tendon path, and the influence of the flexible skin hinder precise control. Nevertheless, these results demonstrate that the RIM Hand can reliably follow intended motion for practical use.

# 4 Experiment

## 4.1 Robot Hand Overview

The final configuration of the robotic hand is illustrated in Section III (Fig. 4(d)). A flexible outer skin is applied to enhance realism, giving it a human-like appearance. Having a compact plastic structure, the RIM Hand offers a total of 15 degrees of freedom, weighs 0.34 kg without actuators and 0.74 kg including actuators, and presents a lightweight structure compared to conventional robotic hands. A brief summary comparing the RIM Hand's performance with other state-of-the-art robotic hands is presented in Table 5. Note that the RIM Hand exhibits greater palm deformation than comparable robotic hands. The performance and motion characteristics of the proposed design were tested through additional experiments.

## 4.2 Joint Motion Test

Each tendon mechanism, Flexion Tendon, Lumbrical Tendon, and the thumb CMC tendons was evaluated to analyze finger and thumb movements (see Fig. 7). Tendon actuation produced smooth flexion at the MCP, PIP, and DIP joints, as well as multi-directional movements of the thumb. The resulting motions closely resembled as seen in the human hand.

## 4.3 Similarity Test

### 4.3.1 Flexion Angle Motion Capture Test

A motion capture experiment was performed to validate the biomimetic design of the RIM Hand by comparing its palm deformation to its human counterpart. As shown in Fig. 8(a), a three-camera system (Flex Camera, OptiTrack) tracked metacarpal angles while grasping a 70 mm-diameter baseball. Three reflective markers were attached to the third, fourth, and fifth metacarpals on both the RIM Hand and the human hand, and the measured data were analyzed in MATLAB.

To characterize metacarpal motion, the CMC joint of the third metacarpal was fixed as a reference plane (given Fig. 8(b)). Singular value decomposition (SVD) was applied to the marker coordinates on the fourth and fifth metacarpals to extract orientation eigenvectors and calculate flexion angles relative to this plane. As shown in Fig. 8(c), the human hand exhibited flexion angles of $27^\circ$ and $23^\circ$, for the fourth and fifth metacarpals, respectively, when grasping the baseball. Under the same conditions, the RIM Hand with flexible palm skin showed angles of $27^\circ$ and $29^\circ$, demonstrating the human deformation pattern. This similarity suggests human-like palm flexibility of the RIM Hand, supporting improved the grasping stability and validating its biomimetic design.

### 4.3.2 Palm Deformation Test

To validate palm deformability, a metacarpal compression test was conducted (see Fig. 9(a)). Using a cable attached to a vise, the RIM Hand's metacarpals were horizontally compressed in 3 mm increments up to 18 mm. A maximum force was measured as 32 N by tension-type load cell. The metacarpals deformed into an arched shape with approximately 28% deformation, similar to a human palm (see Fig. 9(b)).

Furthermore, according to Fig. 9(b), compression profiles were nearly identical to their human equivalents (within 1 N), indicating that the mechanical compliance of the RIM Hand closely mimics human characteristics. These results confirm its potential for natural motion and improved grasping stability.

## 4.4 Grasping Stability Test

The RIM Hand's grasping performance was tested using four YCB (Yale-CMU-Berkeley) objects, each representing a distinct grip type: spherical, cylindrical, tripod pinch, and extension. Each actuator was driven with a constant current of 200 mA to perform a simple power grip. for a simple power grip. To assess the effect of palm compliance, the Soft Palm was replaced with a rigid counterpart. As indicated in Fig. 10(a), the Rigid Palm exhibited limited stability, often failing to maintain a secure due to insufficient surface contact. In contrast, Soft Palm consistently achieved stable grasps across all four objects.

### 4.4.1 Payload Test

To verify the anatomical validity of the proposed flexible palm design, a payload test was performed to examine two key factors: increased surface friction from the silicone surface and bending of the fourth and fifth metacarpals. Three palm configurations were investigated: 1) Rigid Palm (no deformation, low friction), 2) Rigid Palm + Friction (rigid structure with a silicone overlay), and 3) Soft Palm (fully silicone-based).

Each configuration gripped a cylindrical can (64 mm diameter, 80 mm height) under a constant 200 mA, with additional weights for measuring maximum payload. As shown in Fig. 10(b), the Rigid Palm supported about 500 g before slipping. Adding silicone friction increased resistance by roughly 100 g. The Soft Palm, combining silicone friction with metacarpal compliance, supported more than twice the payload of the Rigid Palm. Even if the object eventually slipped, the Soft Palm provided notable resistance against gravity, indicating that palm compliance significantly improves grasp stability.

### 4.4.2 Contact Area Test

A final experiment evaluated whether multi-DOF palm deformation improves grasp stability by expanding contact area. Each actuator was driven with 200 mA, and a cylindrical object (65 mm diameter, 120 mm height) was grasped under identical conditions. To compare the Soft Palm and Rigid Palm, a clay layer (about 5 mm thick) mixed with thermochromic paint was applied. The paint changed color when contacts with an object at -18°C or below, enabling both visual and quantitative analysis of contact regions (see Fig. 10(c)).

Excluding the thumb, the Rigid Palm showed a contact area of 4.9 ± 0.2%, while the Soft Palm reached 13.8 ± 0.3%, with a notable increase (more than 3.5 fold) in the regions near the fourth and fifth metacarpals and the thenar area. The deeper color change observed with the Soft Palm indicates more concentrated contact force and improved grasping adaptability. These results confirm that palm flexibility and deformability significantly improve grasp stability and enable more versatile object handling.

# 5 Discussion and Conclusion

In conclusion, this paper presented the RIM Hand, a biomimetic robotic hand that precisely replicates the CMC joints while incorporating Nitinol-supported skeletal structures. Accurate modeling of the complete carpal-to-metacarpal anatomy enables realistic palm deformation, while tendon-driven fingers equipped with Nitinol-based dorsal extensors enhance joint restoration and support skeletal structure. A flexible silicone skin increases contact area and friction, ensuring stable grasps for a variety of objects. Experiments revealed the palm can deform by up to 28% comparable to human-hand flexibility, while payload capacity and contact area notably exceeded those of a rigid-palm design. Motion capture test confirmed structural similarity to the human hand, validating the anthropomorphic features of the RIM Hand.

However, the introduction of a soft palm and compliant material also reveals five critical challenges:

1. **Torque Limitation and Motion Similarity:** Higher torque is required when flexible skin is applied, given the limited motor power. As seen in the Similarity test, identical torque settings result in decreased anthropomorphism with the skin attached.

2. **Mechanical Compliance vs. Structural Robustness:** Enhanced compliance leads to external-impact resilience but may lead to unintended deformations that compromise functionality.

3. **Short life cycle of the Nitinol mechanism:** Current design of the Nitinol wires still has insufficient fatigue life for long-term use on robotic-hand platforms. Therefore, future work should enhance durability by replacing a single thick Nitinol wire with a bundle of several thinner wires that can deliver the same restoring torque while substantially improving fatigue life.

4. **Tendon Slack and Rolling-Joint Instability:** Tendon elongation over repeated operations and potential rolling-joint misalignment introduce control uncertainties, making accurate joint-angle control more difficult.

5. **Dynamic Behavior with Artificial Skin:** The artificial skin adds dy-

namic complexity, especially at larger actuation ranges, requiring more sophisticated control strategies.

To address these challenges, future work will investigate advanced control strategies to compensate for increased nonlinearity and complexity. The discrepancy between the nominal model and the hardware response will be treated as a disturbance and estimated online using recurrent neural networks, such as Gated Recurrent Units or Long Short-Term Memory networks. The estimated signal will be fed back to the low-level controller to cancel nonlinear effects in real time. Combined with complementary model-based estimators, this approach is expected to enhance motion precision while preserving the inherent flexibility of the robotic hand. Additionally, further exploration of optimized tendon-routing mechanisms and structural reinforcements will be necessary to improve robustness without compromising compliance.

Beyond technical improvements, the RIM Hand holds strong potential for prosthetics, humanoid robots, and assistive devices, where both dexterity and anthropomorphic design are essential. Integrating sensory feedback, such as tactile sensors within the flexible skin, could further advance manipulation in real-world environments.

## Acknowledgments

This work was supported by the National Research Foundation of Korea(NRF) grant funded by the Korea government(MSIT)(RS-2025-16070605) and the Nano Material Technology Development Program through the NRF funded by Ministry of Science and ICT(RS-2025-25442536).

# References

[1] Agneta Gustus, Georg Stillfried, Judith Visser, Henrik Jörntell, and Patrick van der Smagt. Human hand modelling: kinematics, dynamics, applications. *Biological cybernetics*, 106:741–755, 2012.

[2] Honghai Liu. Exploring human hand capabilities into embedded multifingered object manipulation. *IEEE Transactions on Industrial Informatics*, 7(3):389–398, 2011.

[3] Otto Bock. Michelangelo operation manual. *Otto Bock: Duderstadt, Germany*, 2012.

[4] ottobock. bebionic hand. https://www.ottobock.com/en-us/product/8E7*. Accessed: 2025-01-27.

[5] ossur. ilimb. https://www.ossur.com/ko-kr/%EB%B3%B4%EC%A1%B0%EA%B8%B0/%EC%95%94/i-limb-access. Accessed: 2025-01-27.

[6] Ravi Balasubramanian and Veronica J Santos. *The human hand as an inspiration for robot hand development*, volume 95. Springer, 2014.

[7] Joseph T Belter, Jacob L Segil, and BS Sm. Mechanical design and performance specifications of anthropomorphic prosthetic hands: a review. *Journal of rehabilitation research and development*, 50(5):599, 2013.

[8] M Vande Weghe, Matthew Rogers, Michael Weissert, and Yoky Matsuoka. The act hand: Design of the skeletal structure. In *IEEE International Conference on Robotics and Automation, 2004. Proceedings. ICRA'04. 2004*, volume 4, pages 3375–3379. IEEE, 2004.

[9] Larisa Dunai, Martin Novak, and Carmen Garc´ıa Espert. Human hand anatomy-based prosthetic hand. *Sensors*, 21(1):137, 2020.

[10] Dhruv Sharma, Kirti Tokas, Aviral Puri, and Krishna Sharda. Shadow hand. *Journal of Advance Research in Applied Science ISSN*, 2208:2352, 2014.

[11] Markus Grebenstein, Maxime Chalon, Werner Friedl, Sami Haddadin, Thomas Wimböck, Gerd Hirzinger, and Roland Siegwart. The hand of the dlr hand arm system: Designed for interaction. *The International Journal of Robotics Research*, 31(13):1531–1555, 2012.

[12] Pascal Weiner, Julia Starke, Felix Hundhausen, Jonas Beil, and Tamim Asfour. The kit prosthetic hand: design and control. In *2018 IEEE/RSJ International Conference on Intelligent Robots and Systems (IROS)*, pages 3328–3334. IEEE, 2018.

[13] Matteo Laffranchi, Nicolo Boccardo, Simone Traverso, Lorenzo Lombardi, Michele Canepa, Andrea Lince, Marianna Semprini, Jody A Saglia, Abdeldjallil Naceri, Rinaldo Sacchetti, et al. The hannes hand prosthesis replicates the key biological properties of the human hand. *Science robotics*, 5(46):eabb0467, 2020.

[14] Huan Liu, Kai Xu, Bruno Siciliano, and Fanny Ficuciello. The mero hand: A mechanically robust anthropomorphic prosthetic hand using novel compliant rolling contact joint. In *2019 IEEE/ASME International Conference on Advanced Intelligent Mechatronics (AIM)*, pages 126–132. IEEE, 2019.

[15] Yong-Jae Kim, Junsuk Yoon, and Young-Woo Sim. Fluid lubricated dexterous finger mechanism for human-like impact absorbing capability. *IEEE Robotics and Automation Letters*, 4(4):3971–3978, 2019.

[16] Dai Chu, Caihua Xiong, Zhiyi Huang, Jinhao Yang, Jiaji Ma, Jiarui Zhang, Bai-yang Sun, and Junjie Cai. Human palm performance evaluation and the

palm design of humanoid robotic hands. *IEEE Robotics and Automation Letters*, 2024.

[17] Haosen Yang, Guowu Wei, Lei Ren, Zhihui Qian, Kunyang Wang, Haohua Xiu, and Wei Liang. A low-cost linkage-spring-tendon-integrated compliant anthropomorphic robotic hand: Mcr-hand iii. *Mechanism and Machine Theory*, 158:104210, 2021.

[18] Daniele Leonardis and Antonio Frisoli. Cora hand: a 3d printed robotic hand designed for robustness and compliance. *Meccanica*, 55(8):1623–1638, 2020.

[19] Raymond Ma and Aaron Dollar. Yale openhand project: Optimizing open-source hand designs for ease of fabrication and adoption. *IEEE Robotics & Automation Magazine*, 24(1):32–40, 2017.

[20] James Wamai Mwangi, Linh T Nguyen, Viet D Bui, Thomas Berger, Henning Zeidler, and Andreas Schubert. Nitinol manufacturing and micromachining: A review of processes and their suitability in processing medical-grade Nitinol. *Journal of manufacturing processes*, 38:355–369, 2019.

[21] Yash Chitalia, Seokhwan Jeong, Kent K Yamamoto, Joshua J Chern, and Jaydev P Desai. Modeling and control of a 2-dof meso-scale continuum robotic tool for pediatric neurosurgery. *IEEE Transactions on Robotics*, 37(2):520–531, 2020.

[22] Seokhwan Jeong, Yash Chitalia, and Jaydev P Desai. Design, modeling, and control of a coaxially aligned steerable (coast) guidewire robot. *IEEE Robotics and Automation Letters*, 5(3):4947–4954, 2020.

[23] Seokhwan Jeong, Phillip Tran, and Jaydev P Desai. Integration of self-

sealing suction cups on the flexotendon glove-ii robotic exoskeleton system. *IEEE Robotics and Automation Letters*, 5(2):867–874, 2020.

[24] Phillip Tran, Seokhwan Jeong, Steven L Wolf, and Jaydev P Desai. Patient-specific, voice-controlled, robotic flexotendon glove-ii system for spinal cord injury. *IEEE Robotics and Automation Letters*, 5(2):898–905, 2020.

[25] Maxime Chalon, Markus Grebenstein, Thomas Wimböck, and Gerd Hirzinger. The thumb: Guidelines for a robotic design. In *2010 IEEE/RSJ international conference on intelligent robots and systems*, pages 5886–5893. IEEE, 2010.

[26] sizekorea. https://sizekorea.kr. Accessed : 2025-02-18.

[27] Donald A Neumann et al. Kinesiology of the musculoskeletal system. *St. Louis: Mosby*, pages 25–40, 2002.

[28] Christopher H Wise. *Orthopaedic manual physical therapy: from art to evidence*. FA Davis, 2015.

[29] Dennis L Hart, Susan J Isernhagen, and Leonard N Matheson. Guidelines for functional capacity evaluation of people with medical conditions. *Journal of Orthopaedic & Sports Physical Therapy*, 18(6):682–686, 1993.

[30] WP Cooney 3rd, Michael J Lucca, EY Chao, and RL Linscheid. The kinesiology of the thumb trapeziometacarpal joint. *JBJS*, 63(9):1371–1381, 1981.

[31] Maged El-shennawy, Koji Nakamura, Rita M Patterson, and Steven F Viegas. Three-dimensional kinematic analysis of the second through fifth carpometacarpal joints. *The Journal of hand surgery*, 26(6):1030–1035, 2001.

[32] Taisuke Sugaiwa, Hiroyasu Iwata, and Shigeki Sugano. Shock absorbing skin design for human-symbiotic robot at the worst case collision. In

*Humanoids 2008-8th IEEE-RAS International Conference on Humanoid Robots*, pages 481–486. IEEE, 2008.

[33] John-John Cabibihan, Maria Chiara Carrozza, Paolo Dario, Stephane Pattofatto, Moez Jomaa, and Ahmed Benallal. The uncanny valley and the search for human skin-like materials for a prosthetic fingertip. In *2006 6th ieee-ras international conference on humanoid robots*, pages 474–477. IEEE, 2006.

[34] Hyeonjun Park and Donghan Kim. An open-source anthropomorphic robot hand system: Hri hand. *HardwareX*, 7:e00100, 2020.

[35] Marco Controzzi, Marco D Alonzo, Carlo Peccia, Calogero Maria Oddo, Maria Chiara Carrozza, and Christian Cipriani. Bioinspired fingertip for anthropomorphic robotic hands. *Applied Bionics and Biomechanics*, 11(1-2):25–38, 2014.

[36] Haihang Wang, Fares J Abu-Dakka, Tran Nguyen Le, Ville Kyrki, and He Xu. A novel soft robotic hand design with human-inspired soft palm: Achieving a great diversity of grasps. *IEEE Robotics & Automation Magazine*, 28(2):37–49, 2021.

[37] Jean-Michel Boisclair, Thierry Laliberté, and Cl´ement Gosselin. On the optimal design of underactuated fingers using rolling contact joints. *IEEE Robotics and Automation Letters*, 6(3):4656–4663, 2021.

[38] Junmo Yang, Jeongseok Kim, Donghyun Kim, and Dongwon Yun. Shock resistive flexure-based anthropomorphic hand with enhanced payload. *Soft Robotics*, 9(2):266–279, 2022.

[39] Tetsuya Mouri, Haruhisa Kawasaki, Keisuke Yoshikawa, Jun Takai, and Satoshi Ito. Anthropomorphic robot hand: Gifu hand III. In *Proceed-*

*ings of the International Conference on Control, Automation and Systems (ICCAS)*, pages 1288–1293, 2002.

[40] J. D. Weaver, G. M. Sena, K. I. Aycock, A. Roiko, W. M. Falk, S. Sivan, and B. T. Berg. Rotary bend fatigue of Nitinol to one billion cycles. *Shape Memory and Superelasticity*, 9(1):50–73, 2023.

[41] X. J. Yan, D. Z. Yang, and M. Qi. Rotating–bending fatigue of a laser-welded superelastic NiTi alloy wire. *Materials Characterization*, 57(1):58–63, 2006.

[42] Uikyum Kim, Dawoon Jung, Heeyoen Jeong, Jongwoo Park, Hyun-Mok Jung, Joono Cheong, Hyouk Ryeol Choi, Hyunmin Do, and Chanhun Park. Integrated linkage-driven dexterous anthropomorphic robotic hand. *Nature Communications*, 12(1):7177, 2021.

[43] Hume, Mary C and Gellman, Harris and McKellop, Harry and Brumfield Jr, Robert H. Functional range of motion of the joints of the hand. *The Journal of hand surgery*, 15(2):240-243, 1990.

# Tables

Table 1: Dimensions of RIM Hand

| | TIP-DIP (mm) | DIP-PIP (mm) | PIP-MCP (mm) | (3)/(2) | (2)/(1) |
|---|---|---|---|---|---|
| Index | 19.92 | 24.43 | 45.44 | 1.86 | 1.23 |
| Middle | 20.26 | 27.47 | 47.26 | 1.72 | 1.36 |
| Ring | 19.70 | 25.21 | 42.87 | 1.70 | 1.28 |
| Little | 18.44 | 19.43 | 37.12 | 1.91 | 1.05 |

TIP: Fingertip, DIP: Distal-interphalangeal joint, PIP: Proximal-interphalangeal joint, MCP: Metacarpophalangeal joint

Table 2: CMC Joint of RIM Hand

| **Metacarpal** | **1**$^{st}$ | **2**$^{nd}$**-3**$^{rd}$ | **4**$^{th}$ | **5**$^{th}$ |
|---|---|---|---|---|
| **Joint** | Ball | Fixed | Rolling | Rolling |
| **Human Hand ROM** | FL(53°), AB(42°), RO(17°) | Nonaxial small translation | FL(20°), AB(7°) | FL(28-44°), AB(13°) |
| **RIM Hand ROM** | RO(55°) | Fixed | FL(10°) | FL(28-44°) |

FL: Flexion, AB: Abduction, RO: Axial Rotation

Table 3: Fatigue Life of Nitinol in DIP, PIP, MCP Joint

| **Parameter** | **DIP** | **PIP** | **MCP** |
|---|---|---|---|
| $d$ (mm) | 0.58 | 0.58 | 0.58 |
| $\rho$ (mm) | 15 | 18 | 20 |
| $^*\varepsilon_{a1}$ (%) | 0.65 | 0.86 | 0.81 |
| Fatigue life (cycles) | $5.3 \times 10^4$ | $1.2 \times 10^4$ | $1.8 \times 10^4$ |

Fatigue life was calculated with reference to [40]. $^*\varepsilon_{a1} = \frac{d}{2(\frac{d}{2}+\rho)}$, (Calculated by curvature radii of human's average functional angles from [43]).

Table 4: Stress of Nitinol in DIP, PIP, MCP Joint

| **Parameter** | **DIP** | **PIP** | **MCP** |
|---|---|---|---|
| $^*\varepsilon_{a2}$ (%) | 1.93 | 1.63 | 1.01 |
| *stress* (MPa) | 455 | 370 | 355 |

$^*\varepsilon_{a2} = \frac{d}{2\rho}$ (Calculated by curvature radii of human's average functional angles from [43]).

Table 5: DH-Parameter of RIM Hand Middle Finger

| **Link** | **a** (mm) | $\boldsymbol{\alpha}$ (°) | **d** (mm) | $\boldsymbol{\theta}$ (°) |
|---|---|---|---|---|
| 1 | 0 | 0 | 0 | 0 |
| 2 | 13 | 0 | 0 | 0 |
| 3 | 64.5 | 0 | 0 | $\theta_{MCP}/2$ |
| 4 | 18 | 0 | 0 | $\theta_{MCP}/2$ |
| 5 | 32.261 | 0 | 0 | $\theta_{PIP}/2$ |
| 6 | 12 | 0 | 0 | $\theta_{PIP}/2$ |
| 7 | 17.478 | 0 | 0 | $\theta_{PIP}/3$ |
| 8 | 8 | 0 | 0 | $\theta_{PIP}/3$ |
| 9 | 16.261 | 0 | 0 | 0 |

Table 6: Range of Motion of Finger Joints Compared to Other Robotic Hands

| **Joint** | Human ROM (°) | Prosthetic Hand ROM (°) | RIM Hand ROM (°) |
|---|---|---|---|
| **MCP** | 0–90 | 0–90 | 0–90 |
| **PIP** | 0–100 | 0–90 | 0–90 |
| **DIP** | 0–70 | 0–20 | 0–60 |

Table 7: Specification of the RIM Hand Compared with Other Robotic Hands

| | DOF | Thumb DOF | Compliance | Palm Actuation (DOF) | Palm Deformation | Weight |
|---|---|---|---|---|---|---|
| ****CFH-based [38]** | 12 | 4 | O | X | 0 % | 0.91 kg |
| **ILDA [42]** | 15 | 4 | X | X | 0 % | 1.1 kg |
| **FLLEX [15]** | 15 | 4 | O | X | 0 % | 1.81 kg |
| **Gifu [39]** | 16 | 4 | X | X | 0 % | 1.4 kg |
| **Shadow [10]** | 24 | 4 | X | O(1) | 0 % | 4.3 kg |
| **MCR [17]** | 21 | 4 | O | O(1) | 17 % | 0.5 kg |
| **RIM** | **15** | **5** | **O** | **O(2) | **25 %** | **0.74 kg** |

* The characteristics of this robotic hand were evaluated based on its four fingers. ** Passively deformed by the movement of fourth and fifth metacarpals

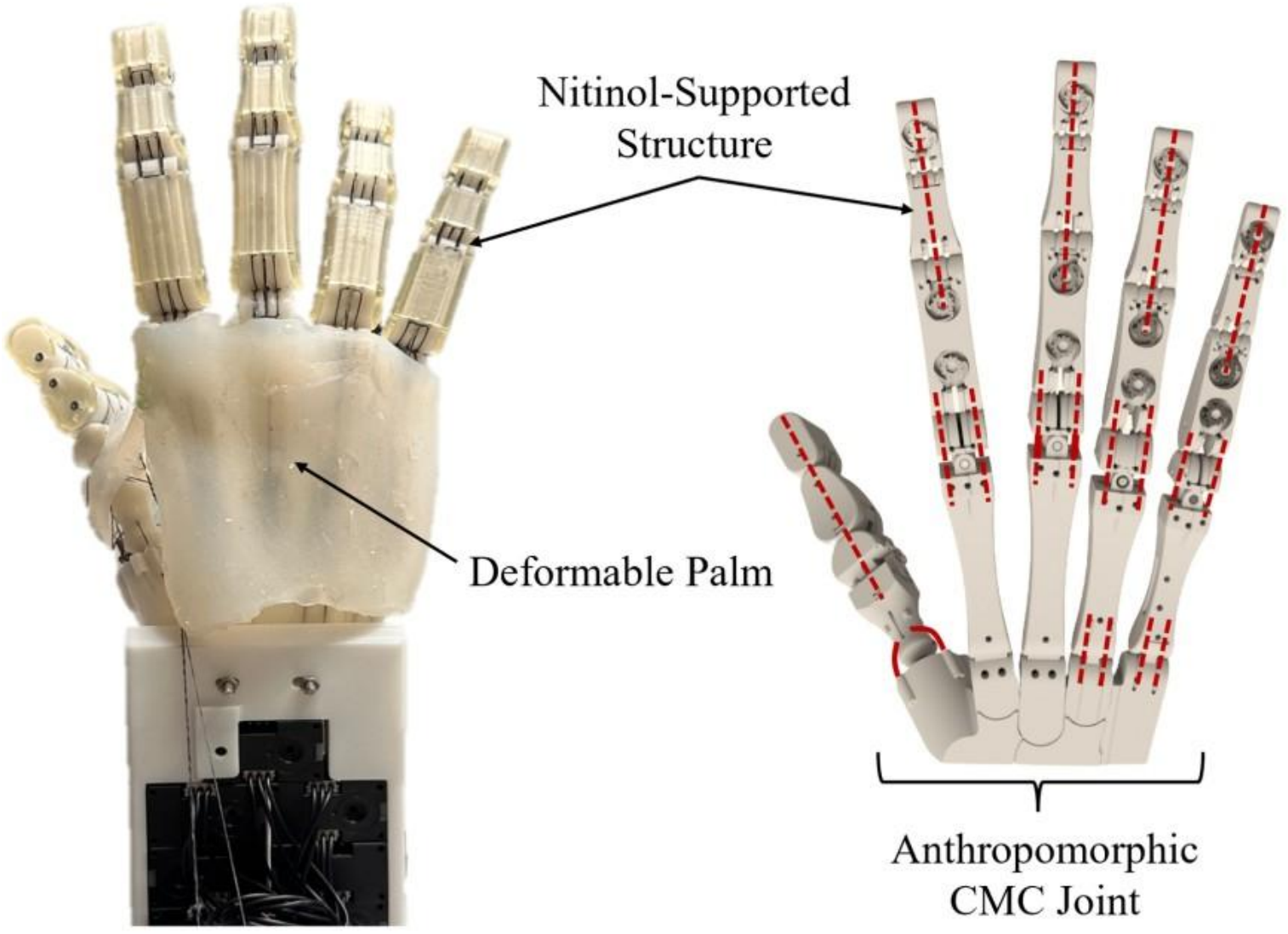

Fig. 1. Overview of the proposed flexible RIM Hand.

195x113mm (300 x 300 DPI)

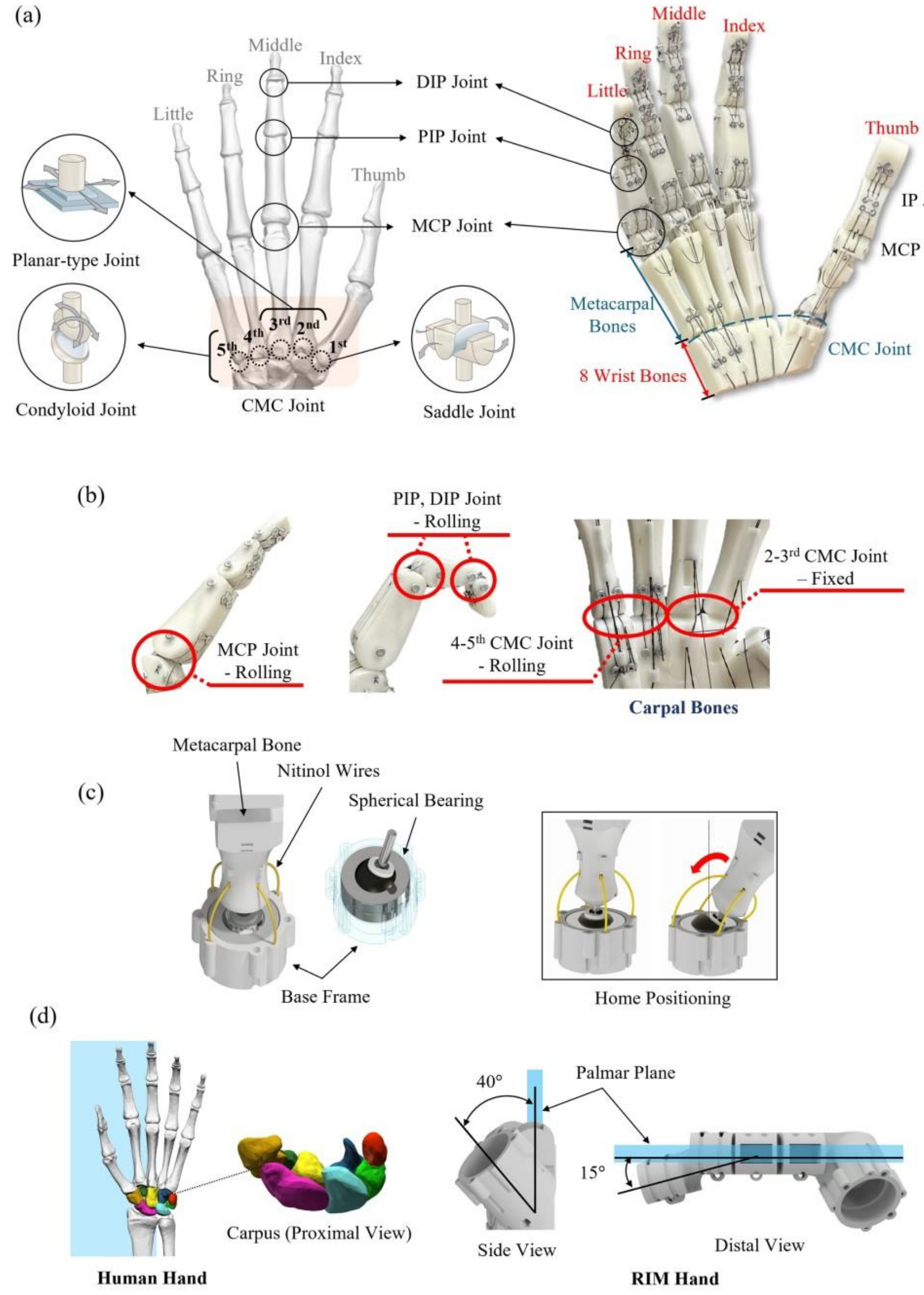

Fig. 2. RIM Hand joint design. (a) Design comparison between human and RIM Hand. (redrawn based on [http://cnx.org/content/col11496/1.6/.]) (b) RIM Hand joint explanation. (c) Ball joint design. (d) Carpal Bones design comparison. (redrawn based on [https://commons.wikimedia.org/wiki/File:Carpus_(left_hand)_10_proximal_view.png], [https://commons.wikimedia.org/wiki/File:Carpus_(left_hand)_11_palmar_view.png])

218x275mm (300 x 300 DPI)

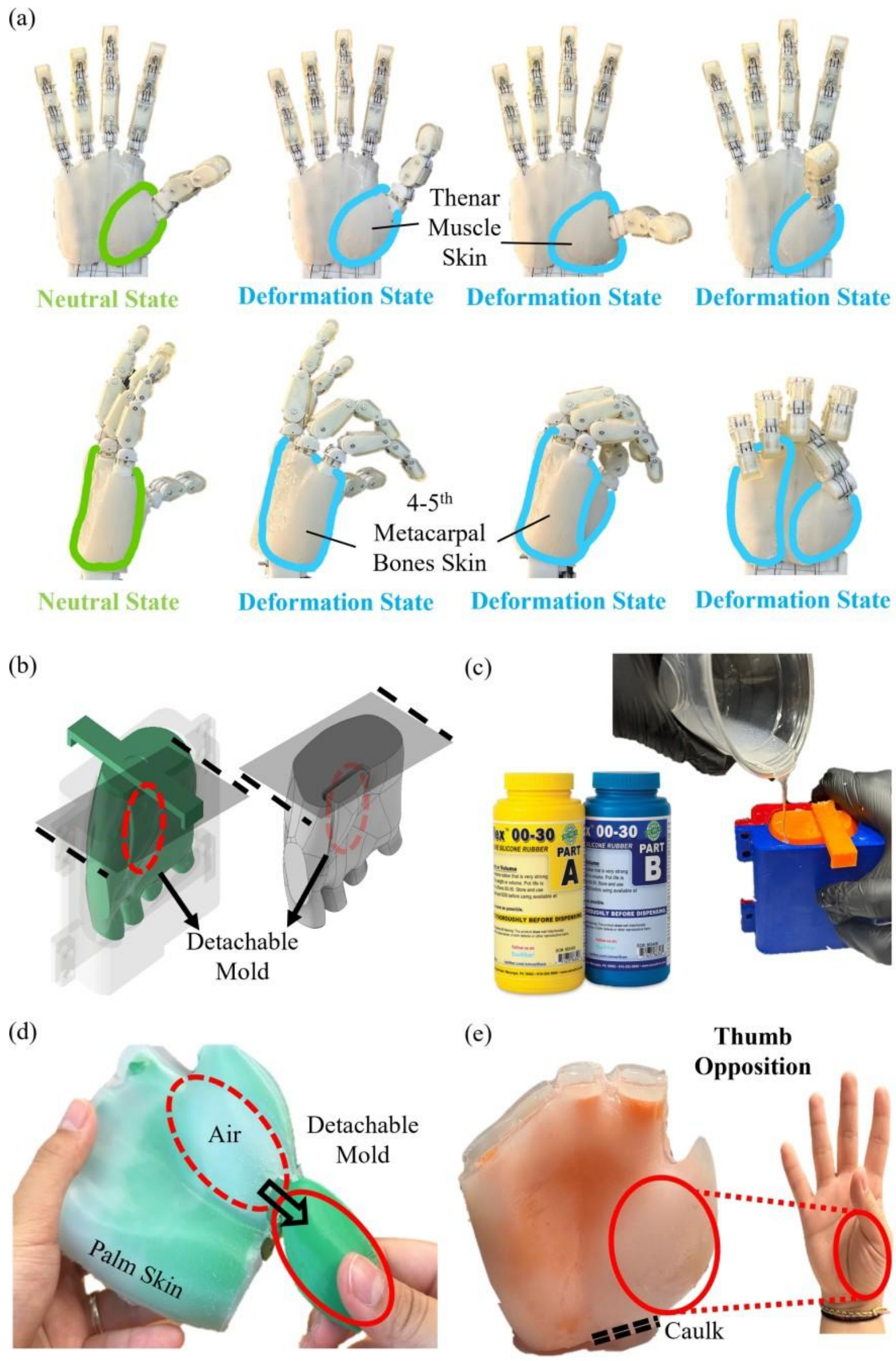

Fig. 3. RIM Hand overall skin design. (a) RIM Hand palm skin deformation with multi-DoF. (b) Palm skin mold 3D printing. (c) Silicone molding. (d) Air pocket generation. (e) Silicone caulking.

190x275mm (300 x 300 DPI)

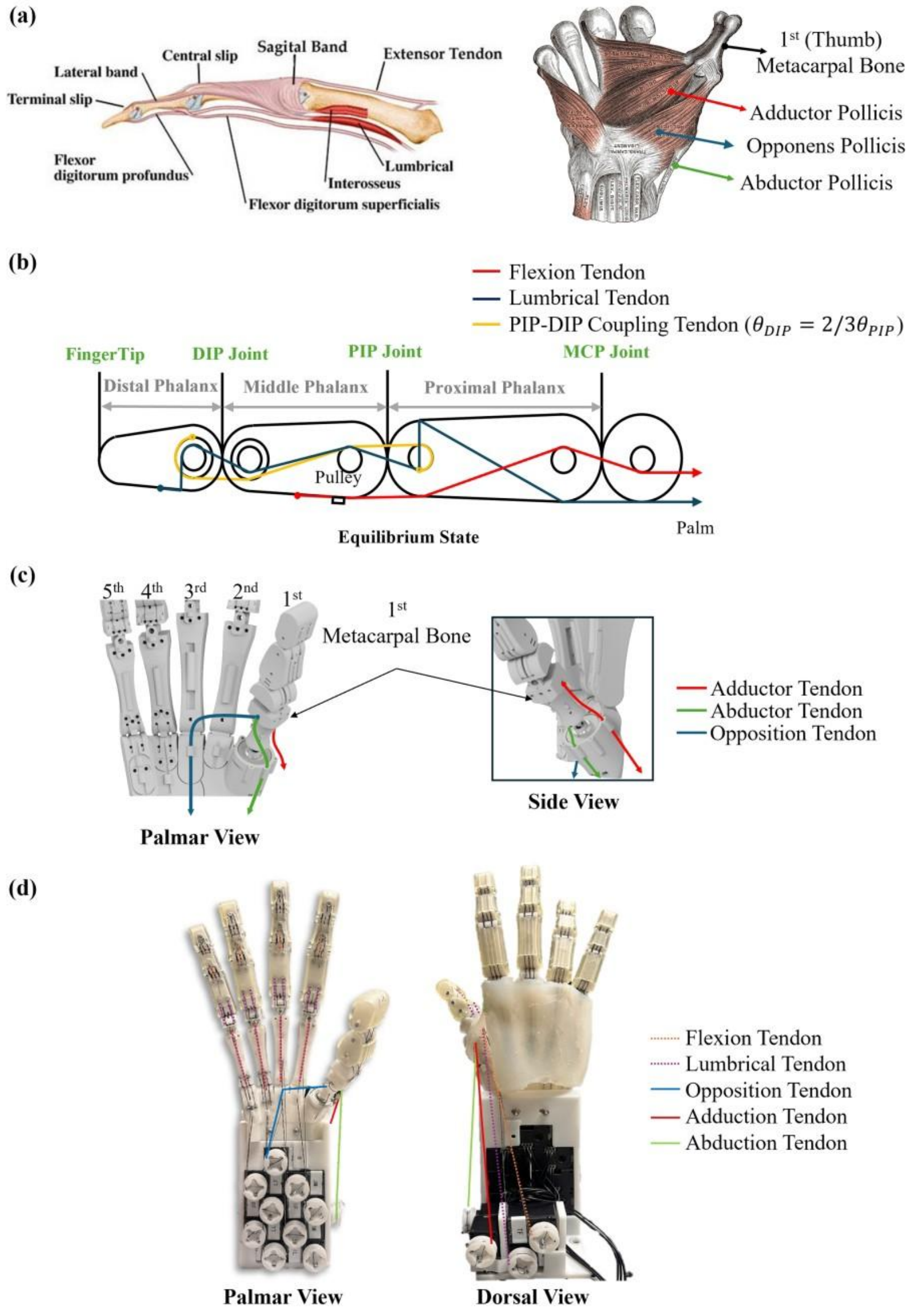

Fig. 4. RIM Hand tendon routing. (a) Tendon routing of human finger and thumb. (redrawn based on [https://fifevirtualhandclinic.co.uk/central-slip-injuries/], [Henry Gray. Anatomy of the human body, volume 8. Lea & Febiger, 1878.]) (b) Tendon routing of RIM Hand MCP, PIP, DIP joint. (c) Tendon routing of RIM Hand thumb. (d) RIM Hand overall tendon routing.

203x289mm (300 x 300 DPI)

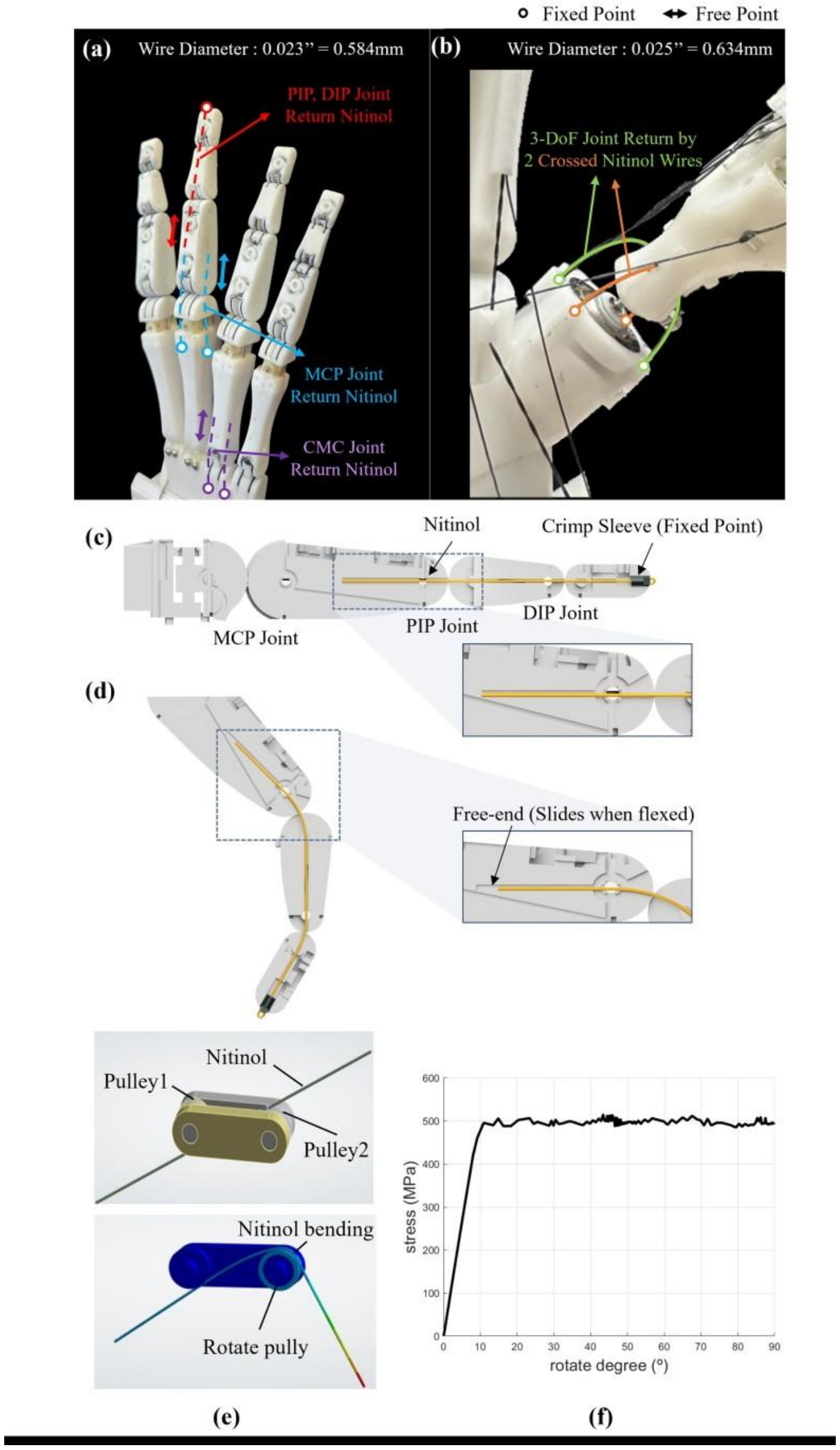

Fig. 5. Nitinol application. (a) Equilibrium state for MCP, PIP, DIP, CMC joint. (b) 1st CMC Joint Return Nitinol Application. (c) Cross-sectional view of middle finger (side view). (d) Cross-sectional view with MCP flexed at 45°, PIP at 45°, and DIP at 30°. (e) ANSYS simulation environment that replicates the Nitinol behavior. (f) Ansys simulation results.

190x332mm (300 x 300 DPI)

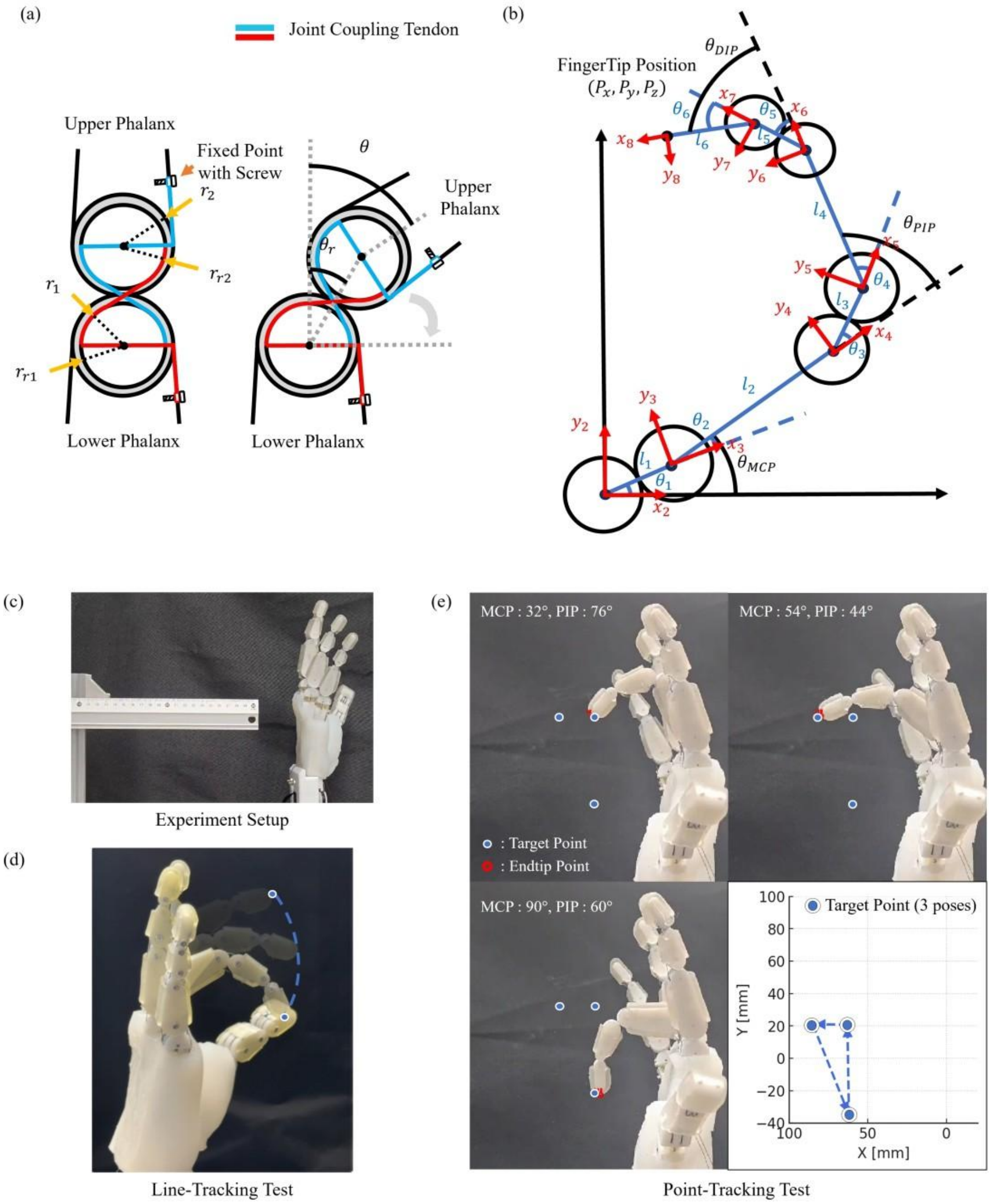

Fig. 6. Finger kinematics. (a) Finger joint kinematics with joint angle. (b) Overall finger kinematics. (c) Experimental setup. (d) Line tracking test. (e) Point tracking test.

273x329mm (300 x 300 DPI)

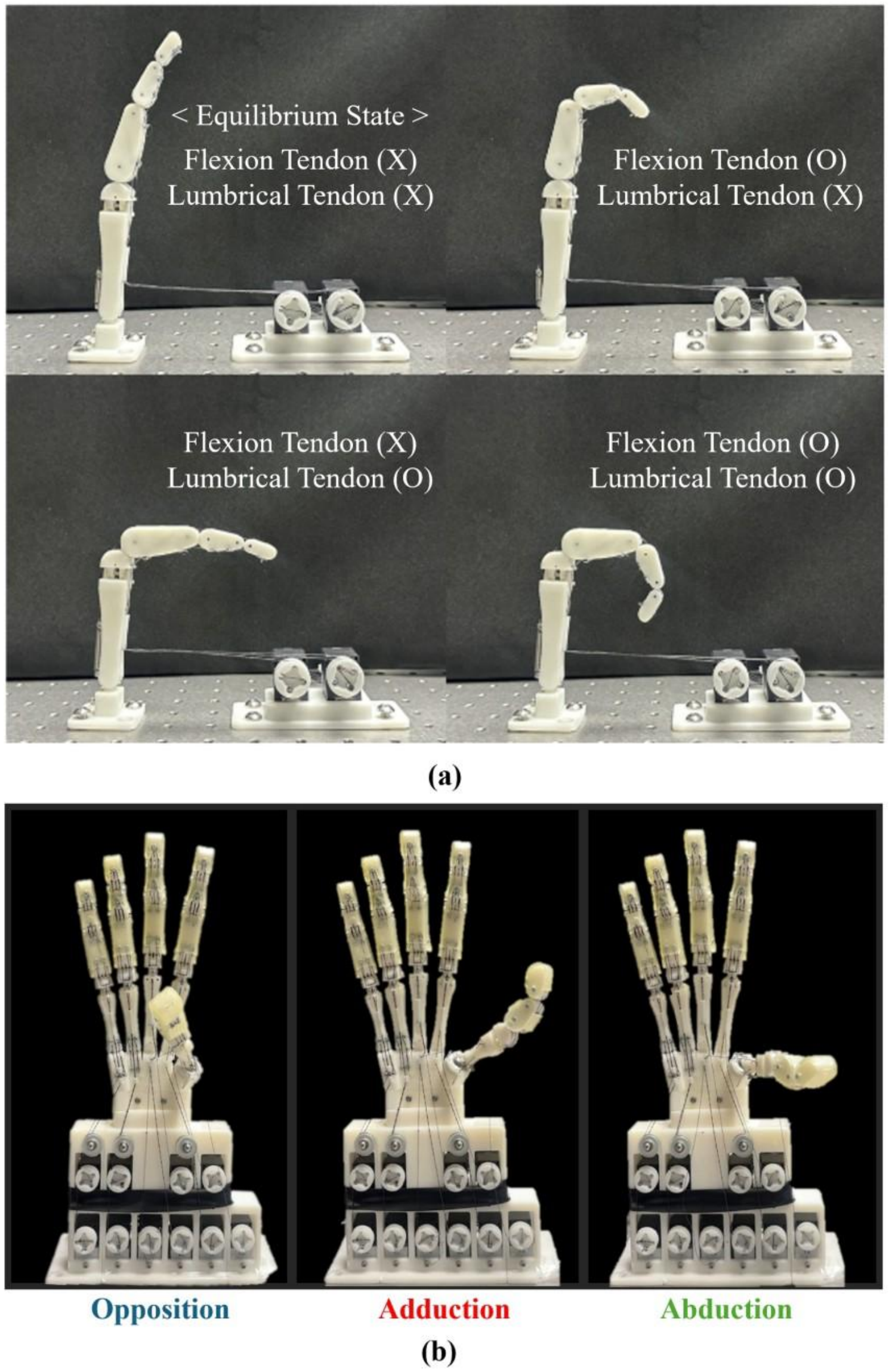

Fig. 7. RIM Hand flexion motion. (a) Finger flexion motion by flexion tendon and lumbrical tendon. (b) Thumb motion by opposition tendon, adduction tendon and abduction tendon.

199x236mm (300 x 300 DPI)

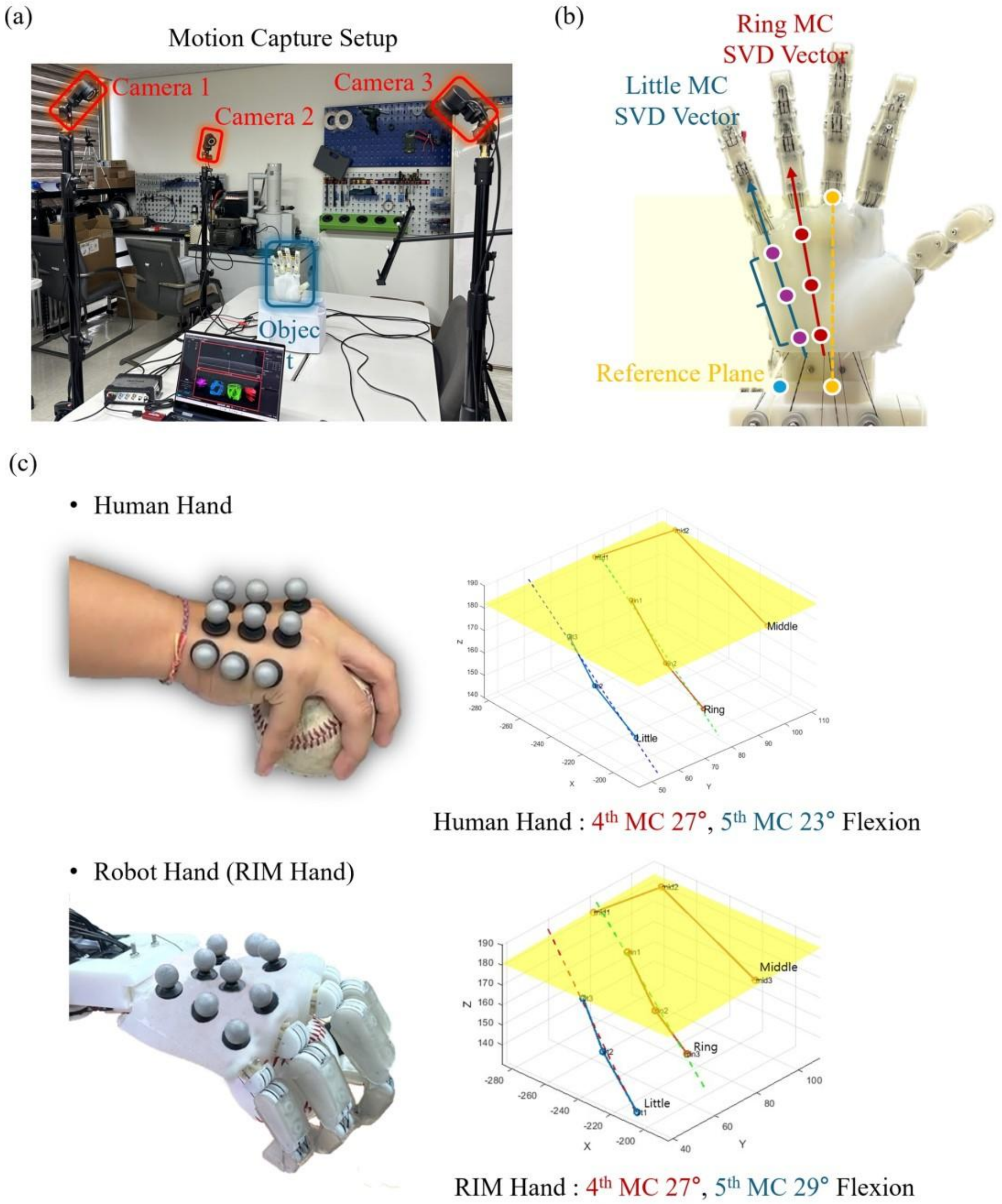

Fig. 8. Similarity test. (a) Test setup with motion capture system. (b) Analysis algorithm of confirming flexion angle of metacarpal bones. (c) RIM Hand similarity test compared with human hand.

200x239mm (300 x 300 DPI)

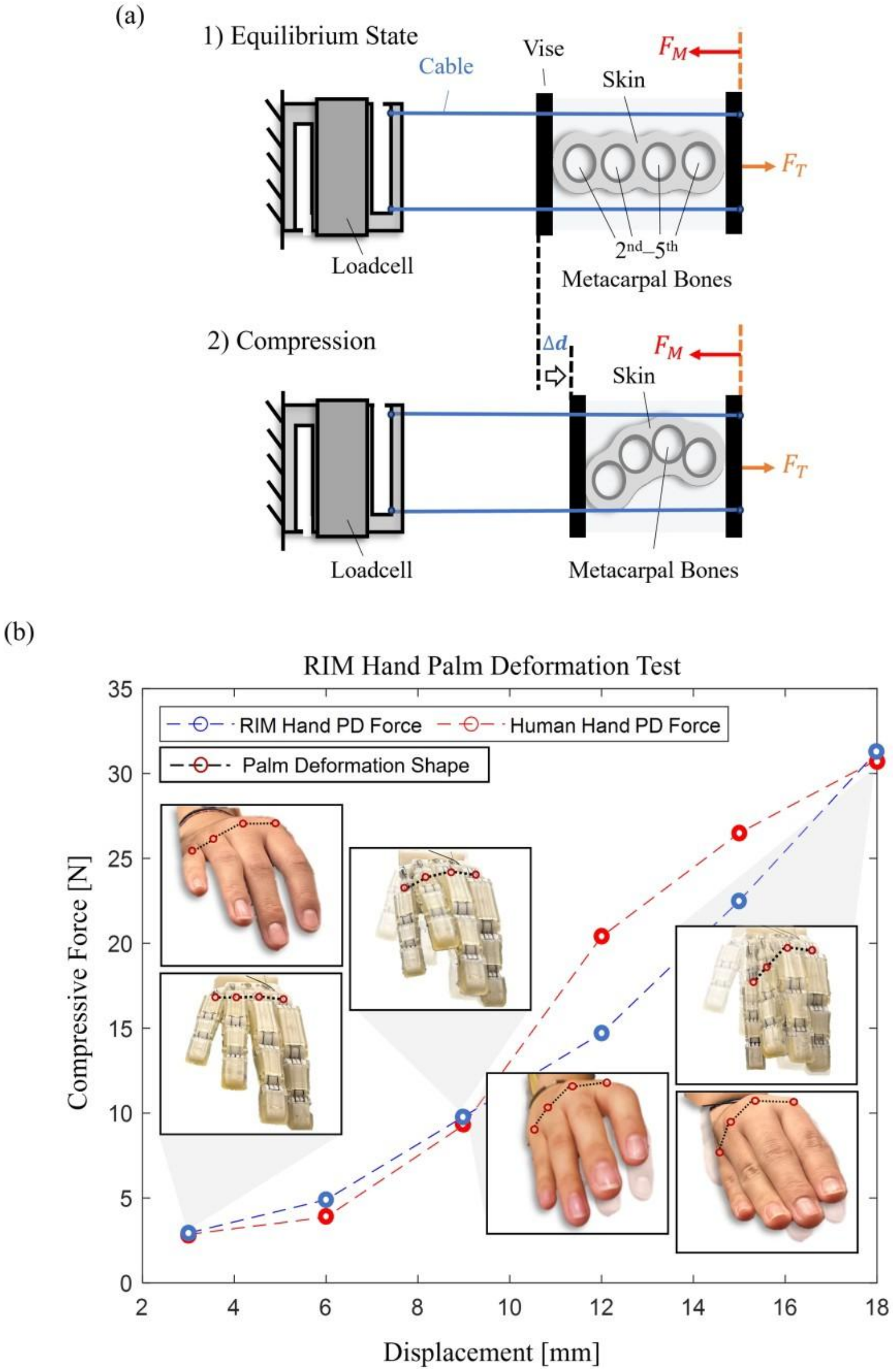

Fig. 9. RIM Hand deformation test. (a) Palm deformation test setup free body diagram. (b) Palm deformation test result. Force–displacement response and the corresponding degree of deformation are presented for both the human hand and the RIM Hand.

205x294mm (300 x 300 DPI)

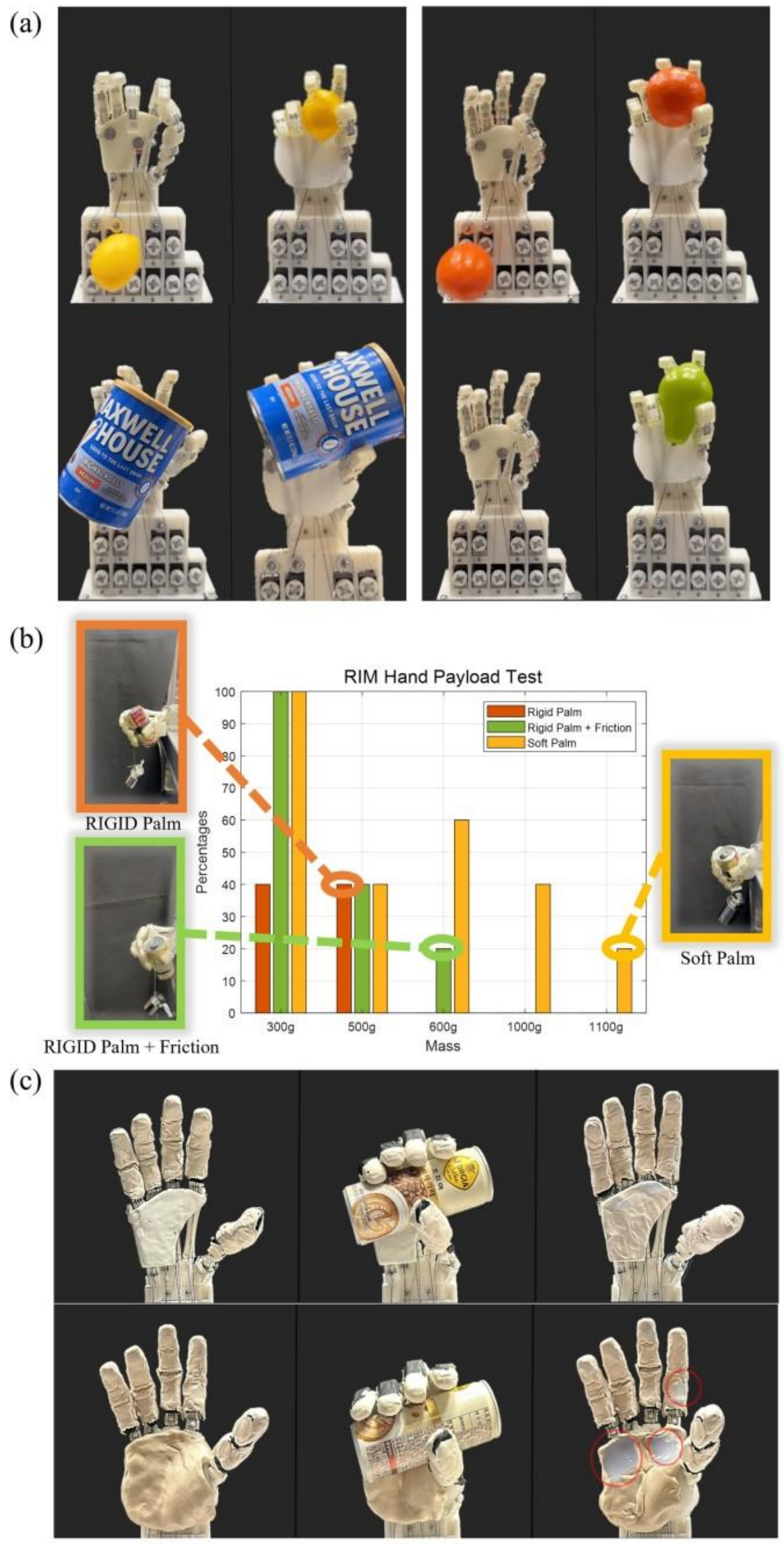

Fig. 10. Test result. (a) RIM Hand grasping test of soft palm compare to rigid palm. (b) RIM Hand payload test compare to three different type of palm skin. (c) RIM Hand object contact area to prove grasping stability.

200x297mm (300 x 300 DPI)